\documentclass[lettersize,journal]{IEEEtran}
\usepackage{tabularx}
\usepackage{soul}
\usepackage{url}
\usepackage[small]{caption}
\usepackage{graphicx}
\usepackage{float}
\usepackage{amsmath,amsfonts}
\usepackage{booktabs}
\usepackage{subfigure}
\usepackage{algorithm}
\usepackage{algorithmic}
\usepackage{amssymb}
\usepackage{enumitem}
\usepackage{amsmath}
\usepackage{multirow}
\usepackage{comment}
\usepackage{makecell}
\usepackage{bm}
\usepackage{color}
\usepackage{colortbl}
\usepackage{amsthm}
\usepackage[html]{xcolor}

\newtheorem{definition}{Definition}
\usepackage{verbatim}
\usepackage{cite}
\usepackage{textcomp}
\usepackage{stfloats}
\usepackage{array} 

\definecolor{sperf}{HTML}{E6EAFD}
\definecolor{bperf}{HTML}{FDEBE7}
\newcommand{\bperf}[1]{\textbf{#1}}
\newcommand{\sperf}[1]{\underline{#1}}

\begin{document}
% \title{Deep Graph Neural Network with Alternating Graph Convolutional and
% Embedding Layers}
\title{ADEdgeDrop: Adversarial Edge Dropping for Robust Graph Neural Networks}

\author{Zhaoliang Chen,
        Zhihao Wu,
        Ylli Sadikaj,
        Claudia Plant,
        Hong-Ning Dai,~\IEEEmembership{Senior Member,~IEEE,}
        Shiping Wang,~\IEEEmembership{Senior Member,~IEEE,}
        Yiu-Ming Cheung,~\IEEEmembership{Fellow,~IEEE,} and
        Wenzhong Guo

\thanks{This work is in part supported by the National Natural Science Foundation of China under Grant U21A20472 and 62276065, and the National Key Research and Development Plan of China under Grant 2021YFB3600503. Corresponding author: Wenzhong Guo.}
\thanks{Zhaoliang Chen is with the College of Computer and Data Science, Fuzhou University, Fuzhou 350116, China, and also with the Department of Computer Science, Hong Kong Baptist University, Hong Kong, China.

Zhihao Wu, Shiping Wang and Wenzhong Guo are with the College of Computer and Data Science, Fuzhou University, Fuzhou 350116, China and also with the Fujian Provincial Key Laboratory of Network Computing and Intelligent Information Processing, Fuzhou University, Fuzhou 350116, China (email: chenzl23@outlook.com, zhihaowu1999@gmail.com, shipingwangphd@163.com, guowenzhong@fzu.edu.cn).

Ylli Sadikaj and Claudia Plant are with the Faculty of Computer Science and with the research network Data Science @ Uni Vienna, University of Vienna, 1090 Vienna, Austria (email: ylli.sadikaj@univie.ac.at, claudia.plant@univie.ac.at).

Hong-Ning Dai and Yiu-Ming Cheung are with the Department of Computer Science, Hong Kong Baptist University, Hong Kong, China (e-mail: henrydai@hkbu.edu.hk, ymc@comp.hkbu.edu.hk).

Zhaoliang Chen and Zhihao Wu contribute equally to this paper.
}
}

\markboth{IEEE Transactions on Knowledge and Data Engineering}%
{Shell \MakeLowercase{\textit{et al.}}: Bare Demo of IEEEtran.cls for IEEE Journals}

\maketitle

\begin{abstract}
    Although Graph Neural Networks (GNNs) have exhibited the powerful ability to gather graph-structured information from neighborhood nodes via various message-passing mechanisms, the performance of GNNs is limited by poor generalization and fragile robustness caused by noisy and redundant graph data. 
    As a prominent solution, Graph Augmentation Learning (GAL) has recently received increasing attention in the literature. 
    Among the existing GAL approaches, edge-dropping methods that randomly remove edges from a graph during training are effective techniques to improve the robustness of GNNs. 
    However, randomly dropping edges often results in bypassing critical edges. 
    Consequently, the effectiveness of message passing is weakened. 
    In this paper, we propose a novel adversarial edge-dropping method (ADEdgeDrop) that leverages an adversarial edge predictor guiding the removal of edges, which can be flexibly incorporated into diverse GNN backbones. 
    Employing an adversarial training framework, the edge predictor utilizes the line graph transformed from the original graph to estimate the edges to be dropped, which improves the interpretability of the edge-dropping method.    
    The proposed ADEdgeDrop is optimized alternately by stochastic gradient descent and projected gradient descent.
    Comprehensive experiments on eight graph benchmark datasets demonstrate that the proposed ADEdgeDrop outperforms state-of-the-art baselines across various GNN backbones, demonstrating improved generalization and robustness.
\end{abstract}
\begin{IEEEkeywords}
Graph neural network, edge dropping, adversarial training, graph augmentation learning, graph representation learning.
\end{IEEEkeywords}

\IEEEpeerreviewmaketitle

\section{Introduction}
As powerful data representations, graphs can represent data items and their complex relations by nodes and edges, respectively. To depict node attributes and rich relational features of edges, diverse Graph Neural Networks (GNNs) have been proposed to extract distinctive node and edge representations~\cite{tu2024attribute,10036442,10328657,ChenDLRGAE22}.
Representative GNN architectures include Graph Convolutional Network (GCN)~\cite{KipfW17}, Graph SAmple and aggreGatE (GraphSAGE)~\cite{HamiltonYL17}, Graph Attention Network (GAT)~\cite{VelickovicCCRLB18}, Simplifying Graph Convolution (SGC)~\cite{WuSZFYW19}, etc. 
During network training, GNNs propagate node features as messages, which are then passed through edges and aggregated~\cite{YangCLNG0HGC22}. 
% As a result, the message-passing mechanism affects the effectiveness of graph learning on top of various GNNs. 
However, inefficient message passing caused by low-quality graph data with noise and redundancy often lets GNNs suffer from overfitting, poor generalization, and less robustness~\cite{mao2022improving,DDingXTL22,chen2020measuring}.

\begin{figure}[!tbp]
  \centering
  \includegraphics[width=0.4\textwidth]{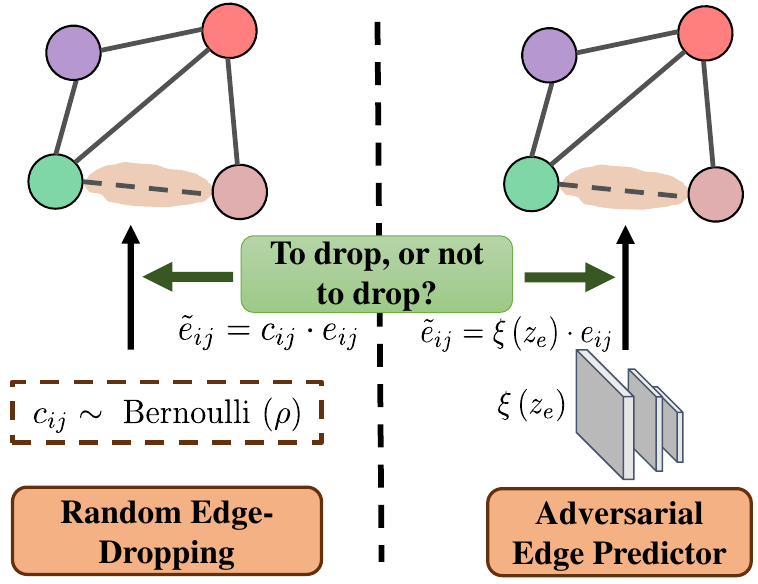}
  \caption{Left (existing methods): random edge dropping with probability $\rho$. Right (our work): edge dropping by an adversarial edge predictor $\xi(\cdot)$.}
  \label{EdgeDrop}     
\end{figure}

To enhance the generalization and robustness of GNN models, Graph Augmentation Learning (GAL) has emerged as a powerful technique for improving node representations through diverse data augmentation methods. 
Different from data augmentation in computer vision, which generates rotated or cropped images to enrich datasets, 
GAL focuses on learning new graph structures or updating node attributes with a diversity of strategies \cite{zhang2024graph,lu2024skipnode}.
One well-known GAL method is utilizing node feature perturbations with random corruption or an adversarial training process, thereby leading to more generalized and robust graph representations~\cite{SureshLHN21,KongLDWZGTG22}.
Another popular approach in GALs is edge dropping, which removes or inhibits redundant edges in a graph during training to promote the robustness of GNNs.
This process is analogous to the dropout technique used for training deep neural networks to mitigate overfitting.
As the most commonly used edge-dropping method, the random edge-dropping strategy~\cite{RongHXH20} removes edges with a certain probability, as illustrated in the left part of Figure~\ref{EdgeDrop}.
However, randomly dropping edges often results in the neglect of critical connections during training as indicated in~\cite{0001XYLWW21,SchlichtkrullCT21,YuanYGJ23}.
%some recent works have pointed out that, owing to the random or uniform edge removals, such a dropping method 
Meanwhile, a graph may include noisy edges or redundant edges, which can be replaced by the integration of other multi-hop paths. 
Therefore, it is crucial to \emph{remove those insignificant edges while retaining those critical edges}, thereby motivating us to design a new edge-dropping approach for GAL.

To cope with the aforementioned issues, we propose a novel edge-dropping method, namely \underline{AD}versarial \underline{Edge} \underline{Drop} (ADEdgeDrop) for arbitrary GNNs. 
Specifically, we design an adversarial edge predictor, which is responsible for determining whether to remove an edge according to the estimated edge embedding trained with a line graph derived from the original graph, as illustrated in the right part of Figure~\ref{EdgeDrop}.
To the best of our knowledge, \textit{this is the first endeavor to design a novel supervised edge-dropping method, which leverages an adversarial edge predictor to effectively remove insignificant edges while preserving critical connections.}
Compared with random edge-dropping methods, the proposed method has the following two advantages. 
(1) \emph{Better semantic interpretability} owing to the edge predictor that evaluates the dropping probability of each edge through the line graph (derived from the original graph), thereby effectively capturing the relationships between the original edges.
This process is supervised by the node attribute homogeneity information that evaluates the probability of correlation between two nodes.
(2) \emph{Improved robustness and generalization} owing to the adversarial training, consequently alleviating the overfitting effect of the edge predictor and the downstream classifier. Different from existing methods that perform adversarial training on node features, the proposed ADEdgeDrop learns a more robust graph structure by the adversarial training on edge embedding in a bottom-up manner.
In summary, the technical contributions of this paper are summarized as follows.
\begin{itemize}
  \item We design an adversarial edge-dropping strategy tailored for arbitrary GNNs. The proposed ADEdgeDrop transforms the original graph into a line graph and introduces trainable perturbations to yield robust edge predictions.
  \item Our proposed ADEdgeDrop can be well incorporated into various GNNs. The edge predictor focuses on the corruption of certain node connections used in the downstream GNNs whose outputs in turn help to update the inputs of the edge predictor.
  \item To fully optimize the permutations and other trainable weights, we propose a joint training algorithm to update these parameters, where a multi-step optimization process is guided by Stochastic Gradient Descent (SGD) and Projected Gradient Descent (PGD).
  \item Comprehensive experiments validate that the proposed method obtains superior performance than existing methods with representative backbone GNNs (GCN, GAT and GraphSAGE) on eight different benchmark datasets by improving the robustness and generalization of the learned graph structure.
\end{itemize}

\section{Related Work}
\subsection{Graph Neural Networks}
As a powerful tool for handling graph-structured data, GNNs have garnered significant attention, 
which play a crucial role in graph representation learning by effectively gathering neighborhood features and propagating vital information across the graph \cite{10152483,10105527,10268633}.
As the well-known example of GNNs, GCNs simplify the graph convolution operator on the non-Euclidean spaces by utilizing a truncated approximation of Chebyshev polynomials \cite{KipfW17}.
Additionally, Velickovic et al. defined a graph attention mechanism to explore the implicit assignment of neighborhood weights for each node \cite{VelickovicCCRLB18}.
Moreover, Hamilton et al. proposed GraphSAGE, a method that learns node embeddings by sampling and accumulating representations from local neighborhood nodes \cite{HamiltonYL17}.
% Leveraging the personalized PageRank algorithm, \citeauthor{KlicperaBG19} improved the propagation schema on graphs \cite{KlicperaBG19}. 
Furthermore, recent research attempts have been made to design simplified GNN variants to enhance GNN computations \cite{WuSZFYW19,ZhuK21,peng2022svd}.
With the rapid development of GNNs, considerable studies have emphasized the significance of graph augmentation to enhance the GNN training and promote the robustness.

\subsection{Graph Augmentation Learning}
GAL aims to enhance the graph structure through data augmentation strategies, which are beneficial in addressing various robustness issues of GNNs \cite{YuHD022,zhang2023spectral,10310152}.
On one hand, some endeavors have adopted adversarial training for graph representation to achieve robust node embeddings~\cite{jin2020graph,kong2022robust,XuDT22}.
On the other hand, graph structure learning methods have attempted to learn richer or less connections from the original sparse graph to augment the training data.
For example, Xu et al. presented a graph rewiring and preprocessing model guided by effective resistance, which can learn reasonable edge addition or removal of a graph \cite{10521752}.
Zhao et al. proposed the GAUG method to explore likely missing edges and remove existing edges by an edge predictor \cite{0003LNW0S21} though GAUG may encounter huge computational costs, particularly when dealing with a large number of nodes. 
As a result, edge-dropping or node-dropping strategies on existing connections are economical approaches to mitigate computational costs since recent studies have demonstrated the promising performance of these dropping methods~\cite{luo2021learning}, such as DropEdge~\cite{RongHXH20,10195874} and DropNode~\cite{FengZDHLXYK020}.
% \cite{RongHXH20,FengZDHLXYK020,luo2021learning}.
% In particular, \citeauthor{RongHXH20} presented the DropEdge method, which randomly removes node connections \cite{RongHXH20}.
% Similarly, \citeauthor{FengZDHLXYK020} proposed the DropNode method, which randomly dropped nodes \cite{FengZDHLXYK020}.
Additionally, Fang et al. devised a general message-dropping method that simultaneously considers various types of dropping strategies \cite{fang2023dropmessage}.
However, these methods suffer from low-quality graphs caused by random removal of nodes or edges, thereby directly influencing the performance of downstream tasks.
Randomly removing edges without any criteria may have a negative impact on the graph representation, particularly if important connections are frequently eliminated during training.
Hence, the key challenge of edge-dropping methods lies in identifying and removing incongruous connections rather than the essential node relationships.

\section{Preliminaries}
\subsection{Notations}
In this paper, we define a graph as $\mathcal{G} = \{ \mathcal{V}, \mathcal{E} \}$, where $\mathcal{V}$ is the set of vertices and $\mathcal{E}$ is the set of edges. The feature matrix and the adjacency matrix of the graph $\mathcal{G}$ are denoted by $\mathbf{X}$ and $\mathbf{A}$, respectively.
Similarly, the line graph transformed from the original graph $\mathcal{G}$ is represented as $\mathcal{G}_{lg} = \{ \mathcal{V}_{lg}, \mathcal{E}_{lg} \}$, whose feature matrix and adjacency matrix are denoted by $\mathbf{X}_{lg}$ and $\mathbf{A}_{lg}$, respectively.
Specifically,  the node attribute $\mathbf{X}_{lg}$ of the line graph $\mathcal{G}_{lg}$ can also be defined as the edge attribute of the graph $\mathcal{G}$, i.e., $\mathbf{X}_{lg} = \mathbf{X}_{e}$.
The learned node embeddings of $\mathcal{G}$ and $\mathcal{G}_{lg}$ are denoted by $\mathbf{Z}$ and $\mathbf{Z}_{lg}$, respectively.
In particular, the $p$-th node in the line graph $\mathcal{G}_{lg}$ corresponds to the $p$-th connective edge between node $i$ and node $j$ in the original graph $\mathcal{G}$.

In the proposed ADEdgeDrop method, we define the networks of the edge predictor and the downstream GNN as $f_{\bm{\omega}}(\cdot)$ and $f_{\bm{\theta}}(\cdot)$, respectively.
The trainable perturbation to the edge embedding is denoted by $\bm{\delta}$.
With our method, we aim to learn the node embedding $\mathbf{Z}$ through the learned incomplete graph $\tilde{\mathcal{G}}$ with the corrupted adjacency matrix $\tilde{\mathbf{A}}$.
Eventually, we conduct the semi-supervised classification with the ground truth denoted by $\mathbf{Y}$.

\subsection{GNN-based Edge Dropping}
The message-passing mechanism for the $k$-th layer of GNN is defined as follows:
\begin{equation}\label{messagepassing}
  h_{i}^{(k+1)} = \varphi^{(k)} \left( h_{i}^{(k)}, \operatorname{Agg}_{j \in \mathcal{N}(i)} \left(\phi^{(k)} \left( h^{(k)}_{i}, h^{(k)}_{j}, e_{ij} \right) \right) \right),
\end{equation}
where $\varphi(\cdot)$ and $\phi(\cdot)$ are any differentiable networks.
Herein, $h_{i}^{(k)}$ represents the feature representation of node $i$ at the $k$-th layer, and $\mathcal{N}(i)$ indicates the set of neighboring nodes connected to node $i$.
The message-passing procedure aggregates the feature information from neighborhood nodes via the connective edges denoted by $e_{ij}$.
The function $\operatorname{Agg}(\cdot)$ defines the aggregation schema and different aggregation functions corresponding to distinct GNN layers.
Specifically, Eq. \eqref{messagepassing} gathers features from the neighbors of node $i$.
Alternatively, we can simplify the GNN architecture to a single-layer representation as
\begin{equation}
  \mathbf{Z} = f_{\bm{\theta}}(\mathbf{X}, \mathbf{A}),
\end{equation}
where $\mathbf{X}$ represents the feature matrix of the graph nodes, and $\mathbf{A}$ is the adjacency matrix encoding all neighborhood connections.
The function $f_{\bm{\theta}}(\cdot)$ is a differentiable function parameterized by the trainable parameter $\bm{\theta}$.

The edge-dropping methods aim to selectively remove connections between nodes to streamline the information aggregation process.
Theoretically, these edge-dropping methods can be summarized as 
\begin{equation}
  \tilde{\mathbf{A}} = \mathbf{C} \odot \mathbf{A},
\end{equation}
where $\tilde{\mathbf{A}}$ is a corrupted adjacency matrix after the edge-dropping process, and $\odot$ denotes the element-wise product operator.
The matrix $\mathbf{C}$ is a corruption matrix that determines whether an edge should be removed or not.
% As a simple example, we can define $\mathbf{C}$ with $\mathbf{C}_{ij} \sim Bernoulli(\rho)$, indicating that $\mathbf{C}_{ij} = 1$ with a probability $\rho$ and vice versa $\mathbf{C}_{ij} = 0$.
As a simple example, we can define $\mathbf{C}$ using a Bernoulli distribution, where each element $\mathbf{C}_{ij} \sim \mathrm{Bernoulli}(\rho)$. 
This means that $\mathbf{C}_{ij} = 1$ with a probability $\rho$, indicating that the edge between nodes $i$ and $j$ should be retained, and $\mathbf{C}_{ij} = 0$ with a probability of $(1-\rho)$ conversely.
\begin{figure*}[!tbp]
    \centering
    \includegraphics[width=0.9\textwidth]{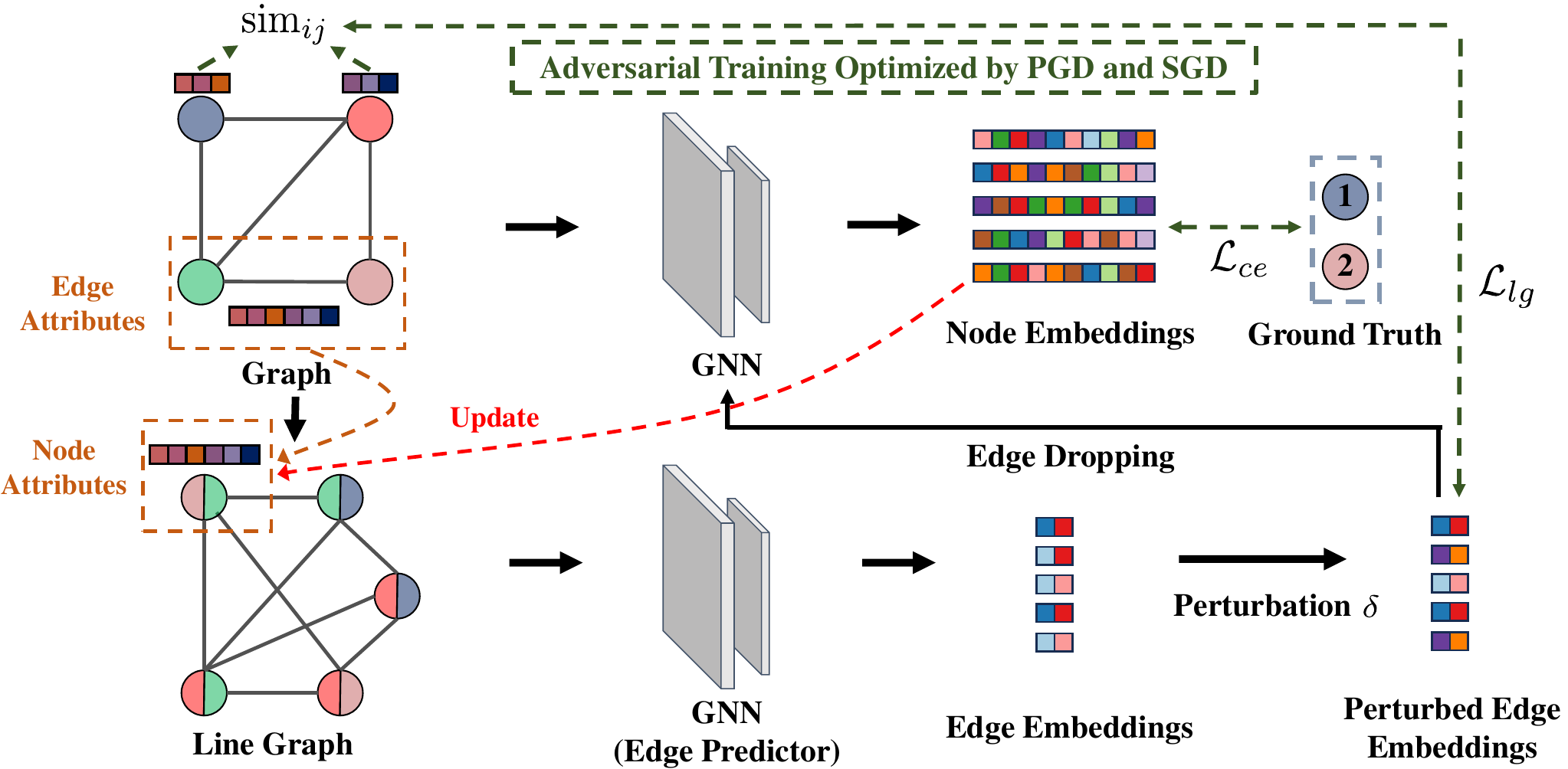}
    \caption{The framework of the proposed ADEdgeDrop, which consists of a basic GNN related to the downstream task, and a GNN-based adversarial edge predictor for the edge-dropping process.}
    \label{Framework}     
  \end{figure*}

  \section{The Proposed Method}
  \subsection{Framework Overview}
  In this section, we elaborate on the proposed ADEdgeDrop method, which tackles the following challenges:
  \begin{itemize}
    \item \textbf{Challenge 1:} \textit{How to construct an adversarial edge predictor for the edge-dropping decision with a line graph transferred from the original graph?}
    \item \textbf{Challenge 2:} \textit{What are the training objectives for the trainable perturbation and other network parameters during the adversarial learning?}
    \item \textbf{Challenge 3:} \textit{How to design a joint training algorithm for both edge predictor and GNN backbone, and how to update all learnable perturbation and parameters?}
  \end{itemize}

  Figure \ref{Framework} demonstrates the overall framework of the proposed ADEdgeDrop, which consists of two primary modules: the general GNN responsible for the downstream tasks and the GNN-based adversarial edge predictor determining the edges to be removed at each iteration.
  The trainable parameters in the framework are updated by a joint training algorithm, which is instructed by PGD and SGD.
  
  \subsection{Robust Edge Dropping}
  First, we solve \textbf{Challenge 1} and elaborate on how to establish the connections between the original graph and the line graph.
  Also, we define the adversarial edge predictor with a trainable perturbation to instruct the edge-dropping process.
  
  \subsubsection{Edge Predictor}
  In pursuit of achieving a robust graph by the robust edge-dropping method, we first transform the initial graph into a line graph $\mathcal{G}_{lg}$ which depicts the attributes of edges in graph $\mathcal{G}$ and relations between them.
  Herein, we give the following definition of the line graph.
  \begin{definition}\label{D1}
    A line graph $\mathcal{G}_{lg} = \{ \mathcal{V}_{lg}, \mathcal{E}_{lg} \}$ describes the relationship between edges in the original graph $\mathcal{G} = \{ \mathcal{V}, \mathcal{E} \}$, i.e., the edge adjacency structure of $\mathcal{G}$. For an undirected graph $\mathcal{G}$, the vertices in the line graph $\mathcal{G}_{lg}$ can be represented as $\mathcal{V}_{lg} = \{ (i \to j): (i, j) \in  \mathcal{E}\}$.
    The edges in $\mathcal{G}_{lg}$ are given by the edge adjacency matrix as follows,
    \begin{equation}
      \left[ \mathbf{A}_{lg} \right]_{(i \to j), (i' \to j')} = 
      \begin{cases}
        1   & j = i', i \neq j',\\
        0   & \mathrm{otherwise}.
    \end{cases}
    \end{equation}
  \end{definition}
  With Definition \ref{D1}, $\left[ \mathbf{A}_{lg} \right]_{pq}$ describes the relationship between the $p$-th node and the $q$-th node in the line graph $\mathcal{G}_{lg}$, corresponding to the connections between the edges $(i \to j)$ and $(i' \to j')$ in the graph $\mathcal{G}$.
  Moreover, we have the definition of node attributes $\mathbf{X}_{lg}$ in the line graph as follows:
  \begin{definition}
  Given a graph $\mathcal{G} = \{ \mathcal{V}, \mathcal{E} \}$ with edge attributes $\mathbf{X}_{e} \in \mathbb{R}^{|\mathcal{E}| \times r}$,
  node attribute $\mathbf{X}_{lg}$ of the line graph $\mathcal{G}_{lg}$ is defined as
  \begin{equation}
    \mathbf{X}_{lg} = \mathbf{X}_{e},
  \end{equation}
  where $[\mathbf{X}_{lg}]_{p} = [\mathbf{X}_{e}]_{p}$ indicates the features of the $p$-th edge $(i \to j)$ in graph $\mathcal{G}$.
  \end{definition}
  
  In contrast to conventional random edge-dropping methods, we consider a supervised edge-dropping strategy guided by an adversarial edge predictor.
  Consequently, the proposed model trains the edge predictor on the line graph $\mathcal{G}_{lg}$ that depicts the features and connections of edges in the original graph $\mathcal{G}$.
  The objective of the edge predictor is to evaluate the necessity of dropping any edges.
  Specifically, the edge predictor aims to retain highly relevant node connections while probabilistically dropping other edges. 
  We define the adversarial edge predictor with a perturbation as
  \begin{equation}\label{edgePrediction}
    \mathbf{Z}_{lg} = f_{\bm{\omega}}(\mathbf{X}_{lg}, \mathbf{A}_{lg})  + \bm{\delta},
  \end{equation}
  where $f_{\bm{\omega}} (\cdot)$ represents any graph neural network, and $\mathbf{Z}_{lg} \in \mathbb{R}^{|\mathcal{V}_{lg}| \times 2}$ is the binary prediction of edges in $\mathcal{G}$.
  % Denoting the edges between node $i$ and $j$,
  % $[\mathbf{Z}_{lg}]_{p1}$ indicates the probability of the edge $p$ exists.
  % Specifically, $[\mathbf{Z}_{lg}]_{p1}$ indicates the probability of the edge $p$ existing between nodes $i$ and $j$ in $\mathcal{G}$.
  Herein, $\bm{\delta}$ is a trainable perturbation, which introduces the corruption to the edge embedding for the adversarial training.
  Therefore, the edge mask $c_{ij}$ for generating a robust and incomplete graph can be derived from the edge prediction, i.e.,
  \begin{equation}\label{edgemask}
    c_{ij} = 
    \begin{cases}
      1   & [\mathbf{Z}_{lg}]_{p1} \geq \mu , \\
      0   & [\mathbf{Z}_{lg}]_{p1} < \mu,
    \end{cases}
  \end{equation}
  where $[\mathbf{Z}_{lg}]_{p1}$ denotes the probability of the $p$-th edge connecting nodes $i$ and $j$ in $\mathcal{G}$.
  The threshold hyperparameter $\mu$ is utilized to determine whether the edge between nodes $i$ and $j$ should be retained.
  With the edge mask, the corrupted adjacency matrix of $\mathcal{G}$ can be represented by
  \begin{equation}\label{maskedEdge}
    \tilde{\mathbf{A}}_{ij} = c_{ij} \odot \mathbf{A}_{ij}.
  \end{equation}
  % Actually, if perturbation $\delta$ is randomly set, the aforementioned edge dropping process is equivalent to the direct random dropout of edges.
  % Nevertheless, we hope to update the perturbation during training, so that a more interpretable and reliable edge dropping process can be developed.
  Note that the edge-dropping process of the ADEdgeDrop method is not equivalent to the random dropout of edges since the perturbation $\bm{\delta}$ is updated during training. 
  This mechanism leads to an interpretable, reliable and robust edge-dropping process, thereby enabling the model to adapt and learn the optimal edge-dropping strategy.

  \subsubsection{GNN with Robust Edge Dropping}
  With the corrupted adjacency matrix $\tilde{\mathbf{A}}$, the GNN-based node embeddings of graph $\mathcal{G}$ can be learned by
  \begin{equation}\label{nodePrediction}
    \mathbf{Z} = f_{\bm{\theta}}(\mathbf{X}, \tilde{\mathbf{A}}),
  \end{equation}
  where $f_{\bm{\theta}}(\cdot)$ is also an arbitrary GNN with parameter $\bm{\theta}$.

  In particular, we update the node attributes in line graph $\mathcal{G}_{lg}$ (i.e., features of edges in $\mathcal{G}$) iteratively, as follows:
  \begin{equation}\label{Xlg}
    [\mathbf{X}_{lg}]_{p}^{(k+1)} = \alpha [\mathbf{X}_{lg}]_{p}^{(k)} + (1 - \alpha) [\hat{\mathbf{X}}_{lg}]_{p}^{(k)},
  \end{equation}
  where $\alpha$ is a trade-off hyperparameter, and the node attributes of the line graph at the $k$-th epoch are updated from the learned GNN embeddings of the original graph, namely,
  \begin{gather}\label{Xlg1}
    [\hat{\mathbf{X}}_{lg}]_{p}^{(k)} = \mathrm{softmax} \left(\operatorname{CONCAT} \left([\mathbf{Z}]_{i}^{(k)},  [\mathbf{Z}]_{j}^{(k)} \right) \right..
  \end{gather}
  Specifically, at the beginning of the training, we can initialize the node features in the line graph with a concatenation of $r(\mathbf{X}_{i})$ and $r(\mathbf{X}_{j})$, where $r(\cdot)$ is any dimensionality reduction method such as PCA or a pretrained autoencoder.
  Subsequently, the model continuously updates $\mathbf{X}_{lg}$ with Eqs. \eqref{Xlg} and \eqref{Xlg1} during training.
  To summarize, the edge predictor decides which edge should be removed, and the basic downstream GNN improves node attributes of the line graph $\mathcal{G}_{lg}$, i.e., input features of the edge predictor.  
  
  \subsubsection{Adversarial Optimization Targets}
  We cope with \textbf{Challenge 2} and elaborate on the targets of the network training.
  For the proposed framework, we need to optimize the training losses of the adversarial edge predictor and the downstream semi-supervised learning task.
  
  First of all, 
  the optimization target of the edge predictor can be formulated as a saddle point problem, i.e., the min-max optimization problem:
  \begin{equation}\label{minmax}
    \min_{\bm{\omega}} \mathbb{E}_{(\mathbf{X}_{lg}, \mathbf{S})\sim \mathcal{D}} \left[ \max_{\|\bm{\delta}\|_{p} \leq \epsilon} \mathcal{L}_{lg} \left(f_{\bm{\omega} }(\mathbf{X}_{lg}, \mathbf{A}_{lg}) + \bm{\delta}, \mathbf{S} \right) \right],
  \end{equation}
  where  $\| \cdot \|_{p}$ denotes the $\ell_{p}$ norm, and $f_{\bm{\omega}} (\cdot)$ is an arbitrary graph neural network parameterized by $\bm{\omega}$.  
  The inner maximization problem aims to find the worst perturbation $\bm{\delta}$ to the outputs of the edge predictor (i.e., the learned edge embeddings of the original graph), and the outer minimization problem seeks the minimal training loss with the vicious noises.
  Consequently, we can learn a robust graph structure with removed connections during training, and utilize it for the downstream GNN model.
  In Eq. \eqref{minmax}, $\mathbf{S}$ is the estimated ground truth of the line graph, which is evaluated from the pair-wise node similarities.
  Thus, it acts as the supervision signal of an edge predictor, which can be formulated as
  \begin{equation}
    \mathbf{S}_{ij} = 
    \begin{cases}
      1   & \mathrm{sim}_{ij} \geq \mu , \\
      0   & \mathrm{sim}_{ij} < \mu,
    \end{cases}
  \end{equation}
  where $\mu$ (the same value as that in Eq. \eqref{edgemask}) is the threshold indicating whether an edge between nodes $i$ and $j$ should be retained during training, and $\mathrm{sim}_{ij}$ is the similarity between two attributed nodes, measured by the Gaussian kernel, i.e., 
%   $\mathrm{sim}_{ij} = \exp \left(- \frac{\| \mathbf{X}_{i} - \mathbf{X}_{j} \|^{2}_{2}}{2\sigma^{2}} \right)$.
  \begin{equation}\label{GaussianSim}
    \mathrm{sim}_{ij} = \exp \left(- \frac{\| \mathbf{X}_{i} - \mathbf{X}_{j} \|^{2}_{2}}{2\sigma^{2}} \right)
  \end{equation}
  
  The loss function $\mathcal{L}_{lg}(\cdot)$ measures the quality of the learned edge embedding, defined as
  \begin{gather}\label{recError}
    \begin{split}
    \mathcal{L}_{lg} &\left( f_{\bm{\omega} }(\mathbf{X}_{lg}, \mathbf{A}_{lg}) + \bm{\delta}, \mathbf{S}  \right) = \\
     &- \frac{1}{\kappa} \sum_{i,j} \mathbf{S}_{ij} \log \left[ \sigma( f_{\bm{\omega} }(\mathbf{X}_{lg}, \mathbf{A}_{lg}) + \bm{\delta} )\right]_{p1},
    \end{split}
  \end{gather}
  where $\sigma(\cdot)$ is an arbitrary activation function and $\kappa$ is the number of non-zero entries in $\mathbf{S}_{ij}$.
  This equation encourages the edge predictor to retain the connections of highly-related nodes at the most and drop some potential unnecessary edges according to the node similarities in $\mathcal{G}$.
  
  Eventually, for the semi-supervised node classification task, the training loss $\mathcal{L}_{ce}$ is defined as follows:
  \begin{gather}\label{celoss}
    \mathcal{L}_{ce} (f_{\bm{\theta}}(\mathbf{X} , \tilde{\mathbf{A}}), \mathbf{Y} ) = 
    - \sum_{i \in \Omega} \sum_{j=1}^{c} \mathbf{Y}_{ij} \mathrm{ln} [f_{\bm{\theta}}(\mathbf{X} , \tilde{\mathbf{A}})]_{ij},
  \end{gather}
  where $\mathbf{Y}$ is the supervision signal and $\Omega$ is the training set.

  \subsection{Training Algorithms}
  We next deal with \textbf{Challenge 3} and provide the joint training strategy for the aforementioned losses.
  The optimization target defined in Eq. \eqref{minmax} can be resolved by alternating updates guided by SGD and PGD methods \cite{MadryMSTV18}.
  SGD aims to address the outer minimization target and PGD handles the inner maximization problem.
  Specifically, the detailed optimization steps are elaborated as follows.
  \begin{algorithm}[!tbp]
    \caption{ADEdgeDrop Training Process}
    \label{algorithm}
    \textbf{Input}: Graph $\mathcal{G}$ with node features $\mathbf{X} \in\mathbb{R}^{n \times m}$ and topological adjacency matrix $\mathbf{A} \in\mathbb{R}^{n \times n}$, ground truth $\mathbf{Y} \in\mathbb{R}^{n \times c}$.\\
    \textbf{Output}: Node embedding $\mathbf{Z}$ and corrupted graph $\tilde{\mathcal{G}}$.
    
    \begin{algorithmic}[1]
    \STATE {Initialize trainable weights $\bm{\theta}$ and $\bm{\omega}$;}
    \STATE {Evaluate the pair-wise node similarities $\mathrm{sim}_{ij}$;}
    \STATE {Obtain the node features $\mathbf{X}_{lg}$ and the adjacency matrix $\mathbf{A}_{lg}$ of the line graph $\mathcal{G}_{lg}$;}
    \WHILE {not converge}
        \STATE {Initialize the perturbation $\bm{\delta}$;}
        \FOR {$t = 1, 2, \cdots, \eta$}
          \STATE {Compute the edge embedding $\mathbf{Z}_{lg}$ with Eq. \eqref{edgePrediction};}
          \STATE {Drop edges in $\mathcal{G}$ with Eqs. \eqref{edgemask} and \eqref{maskedEdge};}
          \STATE {Obtain node embeddings $\mathbf{Z}$ with GNN on the corrupted graph $\tilde{\mathcal{G}}$ with the adjacency matrix $\tilde{\mathbf{A}}$;}
          \STATE {Update the perturbation $\bm{\delta}$ with Eq. \eqref{PGDdelta};}
        \ENDFOR
        \STATE {Update trainable parameters $\bm{\omega}$ with back propagation according to Eq. \eqref{PGDthetalg};}
        \STATE {Update trainable parameters $\bm{\theta}$ with back propagation according to Eq. \eqref{PGDtheta};}
        \STATE {Update node attributes $\mathbf{X}_{lg}$ in $\mathcal{G}_{lg}$ with Eq. \eqref{Xlg};}
    \ENDWHILE
    \RETURN {Node embedding $\mathbf{Z}$ and corrupted graph $\tilde{\mathcal{G}}$.}
    \end{algorithmic}
  \end{algorithm}

  \textbf{Optimize $\bm{\delta}$.} The PGD solution of the inner optimization problem w.r.t. the trainable perturbation $\bm{\delta}$ at the $k$-th iteration can be approximated by
  \begin{gather}\label{PGDdelta}
    \begin{split}
    \bm{\delta}_{t+1} =& \operatorname{proj}_{\|\bm{\delta}\|_{p} \leq \epsilon} \left( \bm{\delta}_{t} + \right.\\ 
    &\left. \gamma \cdot \mathrm{sign} \left( \nabla_{\bm{\delta}} \mathcal{L}_{lg}\left(f_{\bm{\omega}^{(k)}}(\mathbf{X}_{lg}, \mathbf{A}_{lg}) + \bm{\delta}_{t}, \mathbf{S} \right) \right) \right),
    \end{split}
  \end{gather}
  which should loop for $\eta$ times at the $k$-th iteration.
  $\gamma$ is a hyperparameter controlling the updating rate.
  In practical, we select $p = \infty $ for the $\ell_{p}$ norm $\| \bm{\delta} \|_{p}$.
  Thus, $\operatorname{proj}_{\|\bm{\delta}\|_{p} \leq \epsilon}$ projects the updated $\bm{\delta}$ onto the $\epsilon$-ball in the $\ell_{\infty}$ norm.
  
  \textbf{Optimize $\bm{\omega}$.} The outer problem w.r.t. trainable parameter $\bm{\omega}$ in the edge predictor can be optimized by SGD. Namely, 
  \begin{equation}\label{PGDthetalg}
    {\bm{\omega}}^{(k+1)} = \bm{\omega}^{(k)} - \frac{l}{\eta} \sum_{t=1}^{\eta} \nabla_{\bm{\omega}} \mathcal{L}_{lg}\left(f_{\bm{\omega}^{(k)}}(\mathbf{X}_{lg},  \mathbf{A}_{lg}) + \bm{\delta}_{t}, \mathbf{S} \right),
  \end{equation}
  where $l$ is the learning rate.
  This equation accumulates gradients of parameters in the edge predictor during perturbation updating, and these parameters are updated after the optimization of perturbation $\bm{\delta}$ has been conducted for $\eta$ times.
  % In practice, we conduct the backward operation in PyTorch for $\eta$ times, as shown in Algorithm \ref{algorithm}.
  
  \textbf{Optimize $\bm{\theta}$.} The updating rule of the trainable weight $\bm{\theta}$ for the downstream GNN is defined as
  \begin{equation}\label{PGDtheta}
    \bm{\theta}^{(k+1)} = \bm{\theta}^{(k)} - l \nabla_{\bm{\theta}} \mathcal{L}_{ce}\left(f_{\bm{\theta}^{(k)}}(\mathbf{X}, \tilde{\mathbf{A}}), \mathbf{Y} \right).
  \end{equation}
  
  Overall, the training process is summarized in Algorithm~\ref{algorithm}, which is conducted by the alternating updating rules based on the gradients computed by Eqs. \eqref{PGDdelta}, \eqref{PGDthetalg} and \eqref{PGDtheta}.
  
  \subsection{Complexity Analysis}
  Since our method is an extension of backbone GNNs, the time complexity and the memory complexity primarily depend on the architectures of the backbones.
  Herein, we take the applied backbones in our experiments as examples, i.e., GCN~\cite{KipfW17}, GAT~\cite{VelickovicCCRLB18} and GraphSAGE~\cite{HamiltonYL17}.
  Denoting the channel number of input and output node features by $m$ and $c$, respectively,
  the time complexities of the compared backbones in this paper are: $\mathcal{O}(|\mathcal{E}|mc)$ (GCN), $\mathcal{O}(|\mathcal{V}|mc + |\mathcal{E}|c)$ (GAT) and $\mathcal{O} (\prod_{i=1}^{K} \mathcal{S}_{i})$ (GraphSAGE).
  Herein, $\mathcal{S}_{i}$ denotes the neighborhood sample size in GraphSAGE, $|\mathcal{V}|$ is the number of nodes and $|\mathcal{E}|$ is the number of edges.
  
  Although we adopt one additional GNN as an edge predictor during training, the backbone of the edge predictor is the same as the downstream GNN.
  Thus, the time complexity of the edge predictor is generally on par with the downstream backbone.
  Because we transform the original graph $\mathcal{G} = \{ \mathcal{V}, \mathcal{E} \}$ into a line graph $\mathcal{G}_{lg} = \{ \mathcal{V}_{lg}, \mathcal{E}_{lg} \}$,
  the number of nodes in the line graph $\mathcal{G}_{lg}$ is $|\mathcal{E}|$ (i.e., $|\mathcal{V}_{lg}| = |\mathcal{E} |$), 
  and the number of edges $|\mathcal{E}_{lg} |$ in graph $\mathcal{G}_{lg}$ is $\frac{1}{2} \sum_{i=1}^{|\mathcal{V} |} d_{i}^2 - |\mathcal{E}|$.
  Note that we organize the neighborhood information as a sparse tensor in PyTorch, which can significantly save computational resources during training.

  \subsection{Strategy for Edge-Dense Graphs}
  From the previous analyses, it can be found that the time consumption of most edge-dropping methods is tightly related to the number of edges in the graph.
  In order to cope with large-scale datasets, especially graphs with a large number of edges, we further propose a strategy for the edge-dropping strategy with graphs where edges are dense.
  The proposed method conducts a pre-dropping process on the line graph to reduce the computational cost of the edge predictor on line graphs.
  Because we assume that the dropping strategy of important edges that connect semantically similar nodes should be seriously considered, our strategy for edge-dense graphs only constructs line graphs on these critical edges.
  Namely, we conduct the pre-dropping process according to node semantic similarities by
  \begin{equation}
    \hat{\mathbf{A}}_{ij} = \mathbf{M}_{ij} \odot \mathbf{A}_{ij},
  \end{equation}
  where $\mathbf{M}_{ij}$ is the estimated edge mask computed by
    \begin{equation}\label{estimated_edge_mask}
    \mathbf{M}_{ij} = 
    \begin{cases}
      1   & \mathrm{sim}_{ij} \geq p_{pre} , \\
      0   & \mathrm{sim}_{ij} < p_{pre},
    \end{cases}
    \end{equation}
  where $p_{pre}$ is the pre-dropping rate, and $\mathrm{sim}_{ij}$ has been pre-computed by Eq. \eqref{GaussianSim}.
  During GNN training, we also conduct an ordinary edge dropping on the original graph to remove more edges randomly.
  Nevertheless, in contrast to the sole edge-dropping methods, the proposed method preserves critical edges during training with the learnable adversarial edge removal w.r.t. semantic-important node relationships.

  \subsection{Discussion}
  The proposed ADEdgeDrop aims to perform a more robust edge-dropping procedure during GNN training.
  They differ from the existing methods in the following aspects:
  \begin{itemize}
      \item ADEdgeDrop is a \emph{trainable} edge-dropping procedure, which adaptively adjusts the removed edges during different GNN training procedures according to node semantic similarities with an adversarial optimization objective.
      \item Different from existing adversarial-training-based methods (e.g., FLAG \cite{KongLDWZGTG22}) that learn robust node embeddings, the proposed method conducts the adversarial training on the line graphs including the relationships between edges in the original graphs, thereby enabling the \emph{adversarial training w.r.t. edge embeddings} in the original graphs.
      \item Consequently, the proposed method is an adversarial edge-dropping strategy, which can improve the robustness of distinct GNNs and offer insights for future GNN training on a \emph{sparser} graph with fewer edges.
  \end{itemize}

  \section{Experimental Evaluation}
  In this section, we conduct quantitative experiments to explore the following research questions of ADEdgeDrop:
  \begin{enumerate}[leftmargin=*,start=1,label={\bfseries RQ\arabic*:}]
    \item \textbf{(Performance)} \textit{How does the proposed model behave compared with prior GAL methods including random dropping strategies?}
    \item \textbf{(Robustness)} \textit{How many edges are dropped during training, and how does the basic backbone perform with the learned incomplete graph?}
    \item  \textbf{(Effectiveness)} \textit{Does the adversarial training perform effectively compared with traditional training methods?}
  \end{enumerate}
  It is noted that our experiments not only focus on performance and training benefits obtained from the proposed edge-dropping method, but also concentrate on the learned improved graph structure with sparser edges.

  \subsection{Experimental Settings}
  \subsubsection{Datasets}
  We consider eight well-known graph datasets for performance evaluation: BlogCatalog\footnote{https://networkrepository.com/soc-BlogCatalog.php}, Pubmed\footnote{https://github.com/kimiyoung/planetoid}, ACM\footnote{https://github.com/zhumeiqiBUPT/AM-GCN\label{ACM}}, Chameleon\footnote{https://github.com/benedekrozemberczki/MUSAE/}, UAI\textsuperscript{\ref{ACM}}, Actor\footnote{https://github.com/CUAI/Non-Homophily-Large-Scale}, CoraFull\footnote{https://github.com/shchur/gnn-benchmark\#datasets} and Arxiv\footnote{https://ogb.stanford.edu/docs/nodeprop/}.
  The details of the graph datasets used in this paper are listed in Table~\ref{DataDescription}, which summarizes the number of nodes, edges, features, and classes.
  % These datasets can be divided into the following types of graphs: literature citation networks (Citeseer, Pubmed and ACM, website link networks (UAI and Chameleon) and social networks (Actor).
  For the semi-supervised node classification task, we follow \cite{KipfW17} and randomly adopt 20 nodes per class for training, 500 nodes for validation, and 1,000 nodes for testing.
  Specifically, we follow the publicly available split mode for the large-scale Arxiv dataset.
  
  \begin{table*}[t!]
    \centering
    \caption{A brief statistics of all tested graph datasets.}
    \begin{tabular}{lcccccccc}
    \toprule
        Datasets & BlogCatalog & Pubmed & ACM & Chameleon & UAI & Actor & CoraFull & Arxiv \\ \midrule
        \# Nodes & 5,196 & 19,717 & 3,025 & 2,277 & 3,067 & 7,600 & 19,793 & 169,343 \\ 
        \# Edges & 141,738 & 44,324 & 13,128 & 31,371 & 28,308 & 26,659 & 63,421 & 1,157,799 \\
        \# Features & 8,189 & 500 & 1,870 & 2,325 & 4,973 & 932 & 8,710 & 128 \\ 
        \# Classes & 6 & 3 & 3 & 5 & 19 & 5 & 70 & 40 \\ \bottomrule
    \end{tabular}
    \label{DataDescription}
  \end{table*}

  \subsubsection{Compared Baselines and Backbones}
  In this paper, we compare the proposed ADEdgeDrop with the state-of-the-art GAL methods, including edge-dropping, node-dropping, and adversarial training approaches.
  Herein, we present the code links and detailed descriptions of these methods as follows:
  \begin{itemize}
    \item \textbf{DropEdge}\footnote{https://github.com/DropEdge/DropEdge} \cite{RongHXH20} is an edge-dropping method that randomly removes edges from the input graph at each training iteration, thereby promoting the robustness of different GNN backbones.
    \item \textbf{DropNode}\footnote{https://github.com/THUDM/GRAND} \cite{FengZDHLXYK020} randomly samples a binary mask for each node as a perturbation to neighborhood nodes, which only allows each node to aggregate information from a subset of its neighbors.
    \item \textbf{GAUG-O}\footnote{https://github.com/zhao-tong/GAug} \cite{0003LNW0S21} utilizes an autoencoder to effectively encode class-homophilic structure to promote intra-class edges and demote inter-class edges of the graph, which also helps to remove or add some edges in the graph.
    \item \textbf{FLAG}\footnote{https://github.com/devnkong/FLAG} \cite{KongLDWZGTG22} iteratively augments node attributes with a gradient-based adversarial training strategy, which introduces the perturbation into the input node features.
    \item \textbf{DropEdge++}\footnote{https://github.com/hanjq17/DropEdgePlus} \cite{10195874} integrates two structure-aware samplers, i.e., layer-increasingly-dependent sampler and the feature-dependent sampler, to drop edges during GNN training.
    \item \textbf{DropMessage}\footnote{https://github.com/LuckyTiger123/DropMessage} \cite{fang2023dropmessage} directly performs the dropping operation during the message passing mechanism, which can be regarded as a unified framework of most random dropping methods.
  \end{itemize}
  
  In addition, the details and code links of the selected GNN backbones are listed as follows:
  \begin{itemize}
    \item \textbf{GCN}\footnote{https://github.com/tkipf/pygcn} \cite{KipfW17} is a well-known GNN model that simplifies the traditional convolutional operation through conducting the first-order approximation of the Chebyshev polynomial.
    \item \textbf{GAT}\footnote{https://github.com/PetarV-/GAT} \cite{VelickovicCCRLB18} is a GNN framework leveraging the attention mechanism to extract the neighborhood information with the implicit assignment of weights.
    \item \textbf{GraphSAGE}\footnote{https://github.com/williamleif/GraphSAGE} \cite{HamiltonYL17} is a GNN model that explores the node features via sampling and aggregating node representations from the local neighbors of a vertex.
  \end{itemize}

\subsubsection{Implemented Details and Parameter Settings}
The proposed ADEdgeDrop is implemented on a PyTorch platform with the Torch Geometric package.
We run all experiments in this paper on a computer with AMD R9-5900X, RTX 4060Ti 16G GPU, and 32G RAM.
Each algorithm is run 5 times and we record the average accuracy and standard deviation.

There are four primary hyperparameters in our model, i.e., the threshold $\mu$, trade-off parameter $\alpha$, PGD iteration number $\eta$, and PGD updating rate $\gamma$.
In all cases, the threshold $\mu$ ranges in $\{0.5, 0.6, \cdots, 1 \}$.
The PGD updating rate is fixed as $\eta=5$, and the other hyperparameters vary on different datasets.
In our experiments, the values of both $\alpha$ and $\gamma$ are selected from $\{0.1, 0.2, \cdots, 1 \}$.
Note that we keep the consistency between backbone GNNs for the original graph and the line graph.

   \begin{table*}[t!]
    \center
    \caption{Performance (Accuracy\% and Stdandard deviation\%) comparison of various GAL methods for semi-supervised node classification, where the best accuracies are highlighted in \textbf{bold} and the second-best ones are \underline{underlined}. OOT: Out-Of-Time (1 day) error, OOM: Out-Of-Memory (16G) error.}\label{Performance}
    \begin{tabular}{clccccccccc}
    \toprule
        GNN Backbones & GAL Methods & Metrics & BlogCatalog & Pubmed & ACM & Chameleon & UAI & Actor & CoraFull & Arxiv  \\ \midrule
        \multirow{19}{*}{GCN \cite{KipfW17}} & \multirow{2}{*}{Original}  & Accuracy & 70.64  & 77.14  & 87.84  & 48.21  & 56.92  & 20.76  & 56.94  & 69.61    \\
        ~ & ~  & Std  & $\pm$1.44  & $\pm$0.44  & $\pm$0.19  & $\pm$0.26  & $\pm$0.67  & $\pm$1.88  & $\pm$0.77  & $\pm$0.15   \\ \cmidrule{2-11}
        ~ & \multirow{2}{*}{DropEdge \cite{RongHXH20}} & Accuracy & 71.22  & 77.24  & \sperf{88.94}  & 48.34  & 57.82  & 19.94  & 57.99  & 69.83  \\ 
        ~ & ~ & Std  & $\pm$2.12  & $\pm$0.93  & \sperf{$\pm$0.51}  & $\pm$0.94  & $\pm$0.81  & $\pm$0.48  & $\pm$0.24  & $\pm$0.48  \\ \cmidrule{2-11}
        ~ & \multirow{2}{*}{DropNode \cite{FengZDHLXYK020}} & Accuracy & \sperf{72.35}  & 76.92  & 85.90  & 48.40  & 58.22  & 22.18  & 57.89  & 69.77  \\ 
        ~ & ~ & Std & \sperf{$\pm$1.2}  & $\pm$0.50  & $\pm$0.41  & $\pm$0.66  & $\pm$1.74  & $\pm$0.91  & $\pm$0.35 & $\pm$0.29    \\ \cmidrule{2-11}
        ~ & \multirow{2}{*}{GAUG-O \cite{0003LNW0S21}} & Accuracy&  70.89  & \multirow{2}{*}{OOT} & 88.80  & 49.12  & 57.03  & 22.90   &  \multirow{2}{*}{OOM} &  \multirow{2}{*}{OOM}   \\ 
        ~ & ~ & Std & $\pm$0.67  & ~ & $\pm$1.09  & $\pm$1.11  & $\pm$3.32  & $\pm$0.44  & ~  & ~   \\ \cmidrule{2-11}
        ~ & \multirow{2}{*}{FLAG \cite{KongLDWZGTG22}} & Accuracy & 70.78  & 75.08  & 88.28  & 48.10  & 56.74  & 21.42  & \sperf{58.04}  & 69.82 \\ 
        ~ & ~ & Std & $\pm$0.75  & $\pm$0.68  & $\pm$0.35  & $\pm$0.83  & $\pm$1.66  & $\pm$0.52  & \sperf{$\pm$0.33}  & $\pm$0.40  \\ \cmidrule{2-11}
        ~ & \multirow{2}{*}{DropEdge++ \cite{10195874}} & Accuracy & 70.66 & 78.14  & 88.02  & 48.94 & \sperf{58.89}  & \bperf{23.84}  & 57.63 & \sperf{69.89} \\ 
         ~ & ~ & Std & $\pm$3.32 & $\pm$0.74  & $\pm$0.39  & $\pm$1.00  & \sperf{$\pm$0.98}  & \bperf{$\pm$1.22} & $\pm$1.14 & \sperf{$\pm$0.33}   \\ \cmidrule{2-11}
        ~ & \multirow{2}{*}{DropMessage \cite{fang2023dropmessage}} & Accuracy & \multirow{2}{*}{OOM} & \sperf{78.49}  & 87.62  & \sperf{49.22}  & 58.70  & 21.94  & 57.04 & \multirow{2}{*}{OOM} \\ 
         ~ & ~ & Std & ~ & \sperf{$\pm$0.32}  & $\pm$2.13  & \sperf{$\pm$0.94}  & $\pm$1.98  & $\pm$0.95  & $\pm$2.35  & ~   \\ \cmidrule{2-11}
        ~ & \multirow{2}{*}{ADEdgeDrop} & Accuracy & \bperf{73.00}  & \bperf{78.52}  & \bperf{89.26}  & \bperf{50.06}  & \bperf{59.78}  & \sperf{23.04}  & \bperf{58.70}  & \bperf{70.19}\\ 
        ~ & ~ & Std & \bperf{$\pm$0.68}  & \bperf{$\pm$0.16}  & \bperf{$\pm$0.08}  & \bperf{$\pm$0.89}  & \bperf{$\pm$0.81}  & \sperf{$\pm$2.52}  & \bperf{$\pm$0.46}  & \bperf{$\pm$0.05}    \\ \bottomrule
        \multirow{19}{*}{GAT \cite{VelickovicCCRLB18}} & \multirow{2}{*}{Original} & Accuracy & 51.14  & 75.94  & 85.98  & 46.22  & 56.68  & 22.94  & 54.98  & 69.28  \\ 
        ~ & ~ & Std  & $\pm$5.34  & $\pm$1.18  & $\pm$0.50  & $\pm$1.55  & $\pm$1.30  & $\pm$1.63  & $\pm$0.59  & $\pm$0.06    \\ \cmidrule{2-11}
        ~ & \multirow{2}{*}{DropEdge \cite{RongHXH20}} & Accuracy & 51.89  & 76.48  & 86.48  & 46.46  & 56.32  & 21.96  & 55.03  & 69.35  \\ 
        ~ & ~ & Std & $\pm$4.13  & $\pm$0.25  & $\pm$0.42  & $\pm$1.54  & $\pm$0.93  & $\pm$1.12  & $\pm$0.47  & $\pm$0.48  \\ \cmidrule{2-11}
        ~ & \multirow{2}{*}{DropNode \cite{FengZDHLXYK020}} & Accuracy & 52.91  & 76.18  & 85.92  & 46.56  & 48.78  & 22.40  & 55.79  & 69.53  \\ 
        ~ & ~ & Std& $\pm$4.89  & $\pm$0.35  & $\pm$0.54  & $\pm$1.76  & $\pm$7.84  & $\pm$1.87  & $\pm$0.47  & $\pm$0.39  \\ \cmidrule{2-11}
        ~ & \multirow{2}{*}{GAUG-O \cite{0003LNW0S21}} & Accuracy & 51.77  & \multirow{2}{*}{OOT} & \bperf{88.28}  & 47.64  & 56.72  & 23.05  & \multirow{2}{*}{OOM} & \multirow{2}{*}{OOM}  \\ 
        ~ & ~ & Std & $\pm$2.39  & ~ & \bperf{$\pm$0.45}  & $\pm$2.26  & $\pm$5.10  & $\pm$1.34  & ~ & ~  \\ \cmidrule{2-11}
        ~ & \multirow{2}{*}{FLAG \cite{KongLDWZGTG22}} & Accuracy & 48.44  & 76.30  & 87.16  & 48.08  & 58.04  & 21.80  & 57.38  & \bperf{69.76}    \\ 
        ~ & ~ & Std  & $\pm$1.34  & $\pm$0.46  & $\pm$0.66  & $\pm$1.27  & $\pm$0.57  & $\pm$0.72  & $\pm$0.54  & \bperf{$\pm$0.31}  \\ \cmidrule{2-11}
        ~ & \multirow{2}{*}{DropEdge++ \cite{10195874}} & Accuracy & \sperf{53.22} & 76.72  & 87.21  & 47.56  & 57.87  &  23.69 & 57.34 & 69.63 \\ 
         ~ & ~ & Std & \sperf{$\pm$1.64} & $\pm$0.76  & $\pm$0.98 & $\pm$0.73  & $\pm$1.43  & $\pm$1.23 & $\pm$0.55 & $\pm$0.22   \\ \cmidrule{2-11}
        ~ & \multirow{2}{*}{DropMessage \cite{fang2023dropmessage}} & Accuracy & \multirow{2}{*}{OOM} & \bperf{77.14}  & 86.42  & \sperf{48.44}  & \sperf{58.20}  & \sperf{23.78}  & \bperf{58.04}  & \multirow{2}{*}{OOM}  \\
        ~ & ~ & Std & ~ & \bperf{$\pm$0.46}  & $\pm$1.15  & \sperf{$\pm$0.90}  & \sperf{$\pm$1.06}  & \sperf{$\pm$1.54}  & \bperf{$\pm$0.62}  & ~ \\ \cmidrule{2-11}
        ~ & \multirow{2}{*}{ADEdgeDrop} & Accuracy & \bperf{54.52}  & \sperf{76.86}  & \sperf{87.86}  & \bperf{48.60}  & \bperf{58.58}  & \bperf{25.60}  & \sperf{57.70}  & \sperf{69.70}   \\
        ~ & ~ & Std & \bperf{$\pm$2.80}  & \sperf{$\pm$1.03}  & \sperf{$\pm$0.52}  & \bperf{$\pm$1.19}  & \bperf{$\pm$0.80}  & \bperf{$\pm$1.37}  & \sperf{$\pm$0.41}  & \sperf{$\pm$0.10}   \\ \bottomrule
        \multirow{19}{*}{GraphSAGE \cite{HamiltonYL17}} & \multirow{2}{*}{Original} & Accuracy & 75.30  & 77.30  & 86.16  & 47.12  & 55.20  & 23.04  & 57.34  & 69.28    \\ 
        ~ & ~ & Std & $\pm$2.87  & $\pm$0.81  & $\pm$0.47  & $\pm$1.70  & $\pm$0.95  & $\pm$1.63  & $\pm$0.67  & $\pm$0.11    \\ \cmidrule{2-11}
        ~ & \multirow{2}{*}{DropEdge \cite{RongHXH20}} & Accuracy & 76.43  & 76.98  & 87.52  & \sperf{48.54}  & 57.48  & 23.22  & \sperf{58.79}  & 69.44  \\ 
        ~ & ~ & Std& $\pm$3.47  & $\pm$1.95  & $\pm$0.25  & \sperf{$\pm$0.99}  & $\pm$0.84  & $\pm$0.62  & \sperf{$\pm$0.72}  & $\pm$0.20  \\ \cmidrule{2-11}
        ~ & \multirow{2}{*}{DropNode \cite{FengZDHLXYK020}} & Accuracy & 77.02  & 76.28  & 87.62  & 47.96  & 56.94  & 21.86  & 57.98  & \sperf{69.73}  \\ 
        ~ & ~ & Std & $\pm$2.82  & $\pm$1.76  & $\pm$0.43  & $\pm$1.04  & $\pm$1.92  & $\pm$0.45  & $\pm$0.44  & \sperf{$\pm$0.29}  \\ \cmidrule{2-11}
        ~ & \multirow{2}{*}{GAUG-O \cite{0003LNW0S21}} & Accuracy& 77.45  & \multirow{2}{*}{OOT} & \sperf{88.10}  & 47.36  & 55.77  & \sperf{23.76}  & \multirow{2}{*}{OOM} & \multirow{2}{*}{OOM} \\
        ~ & ~ & Std & $\pm$3.50  &  ~ & \sperf{$\pm$0.33}  & $\pm$0.83  & $\pm$6.04  & \sperf{$\pm$1.04}  &  ~ &  ~ \\ \cmidrule{2-11}
        ~ & \multirow{2}{*}{FLAG \cite{KongLDWZGTG22}} & Accuracy & 75.70  & 74.04  & 86.14  & 48.12  & 57.48  & 22.22  & 57.84  & 69.62  \\ 
        ~ & ~ & Std & $\pm$0.88  & $\pm$1.29  & $\pm$0.76  & $\pm$1.08  & $\pm$1.67  & $\pm$0.98  & $\pm$0.50  & $\pm$0.16 \\ \cmidrule{2-11}
        ~ & \multirow{2}{*}{DropEdge++ \cite{10195874}} & Accuracy & \sperf{78.61} & 77.52  & 87.93  & 48.31 & 58.11  &  23.49 & 58.71 & 69.58 \\ 
         ~ & ~ & Std & \sperf{$\pm$1.54} &  $\pm$0.44 & $\pm$0.67 & $\pm$0.65  &  $\pm$1.98 & $\pm$2.11 & $\pm$0.85 & $\pm$0.38   \\ \cmidrule{2-11}
        ~ & \multirow{2}{*}{DropMessage \cite{fang2023dropmessage}} & Accuracy & \multirow{2}{*}{OOM} & \sperf{78.08}  & 86.48  & 47.12  & \sperf{59.40}  & 21.76  & 57.58  & \multirow{2}{*}{OOM} \\ 
        ~ & ~ & Std & ~ & \sperf{$\pm$0.39}  & $\pm$0.15  & $\pm$0.53  & \sperf{$\pm$1.74}  & $\pm$2.37  & $\pm$2.56  & ~    \\ \cmidrule{2-11}
        ~ & \multirow{2}{*}{ADEdgeDrop} & Accuracy & \bperf{80.30}  & \bperf{78.12}  & \bperf{88.36}  & \bperf{48.68}  & \bperf{62.98}  & \bperf{25.06}  & \bperf{58.98}  & \bperf{70.35}    \\
        ~ & ~ & Std & \bperf{$\pm$2.05}  & \bperf{$\pm$0.65}  & \bperf{$\pm$0.41}  & \bperf{$\pm$0.93}  & \bperf{$\pm$1.21}  & \bperf{$\pm$2.25}  & \bperf{$\pm$0.62}  & \bperf{$\pm$0.12}\\ \bottomrule
    \end{tabular}
  \end{table*}

\subsection{Experimental Results}

\subsubsection{Performance with Different Backbones (RQ1)}
First, we evaluate the proposed ADEdgeDrop compared with representative baselines under different GNN backbones.
The learning rates of all models are fixed as 0.01 and the Adam optimizer is adopted.
For all GNN backbones except those with DropEdge++, we use a 2-layer network architecture and fix the hidden dimension as 16 on all datasets except Arxiv.
We fix the hidden dimension of backbones on Arxiv dataset as 256 for all models.
Moreover, we also adopt the pre-dropping process on the edge-dense Arxiv dataset.
As reported in Table \ref{Performance}, ADEdgeDrop obtains superior classification accuracy than other compared methods in most cases.
Notably, to achieve desired performance, we need to fix the dimension of hidden layers as 128 or even more for DropEdge++ on BlogCatalog and UAI datasets, which are much larger than all other methods.
This indicates DropEdge++ requires more computational resources to achieve comparable accuracy.
In addition, GAUG-O which also adopts an edge predictor exhibits competitive performance compared with other GAL methods.
Nevertheless, it is time-consuming when addressing dense graphs with more edges, and even cannot obtain the results in an acceptable time on Pubmed, CoraFull and Arxiv datasets.
Compared with FLAG, which conducts adversarial training on features, the proposed ADEdgeDrop generally shows promoting performance.
This observation indicates that adversarial edge learning improves GNN training more significantly.
In conclusion, the experimental results indicate that ADEdgeDrop positively impacts robust representation learning compared with other state-of-the-art GAL methods, especially for random dropping methods like DropEdge, DropNode, DropEdge++ and DropMessage.
% In conclusion, the experimental results display that the proposed ADEdgeDrop boosts the accuracy of backbone GNNs, and has superior performance compared with other state-of-the-art GAL methods, especially other random dropping methods.

\subsubsection{Performance with Edge-Removal and Edge-Addition Attacks (RQ2)}
\begin{figure*}[t!]
    \centering
    \includegraphics[width=\textwidth]{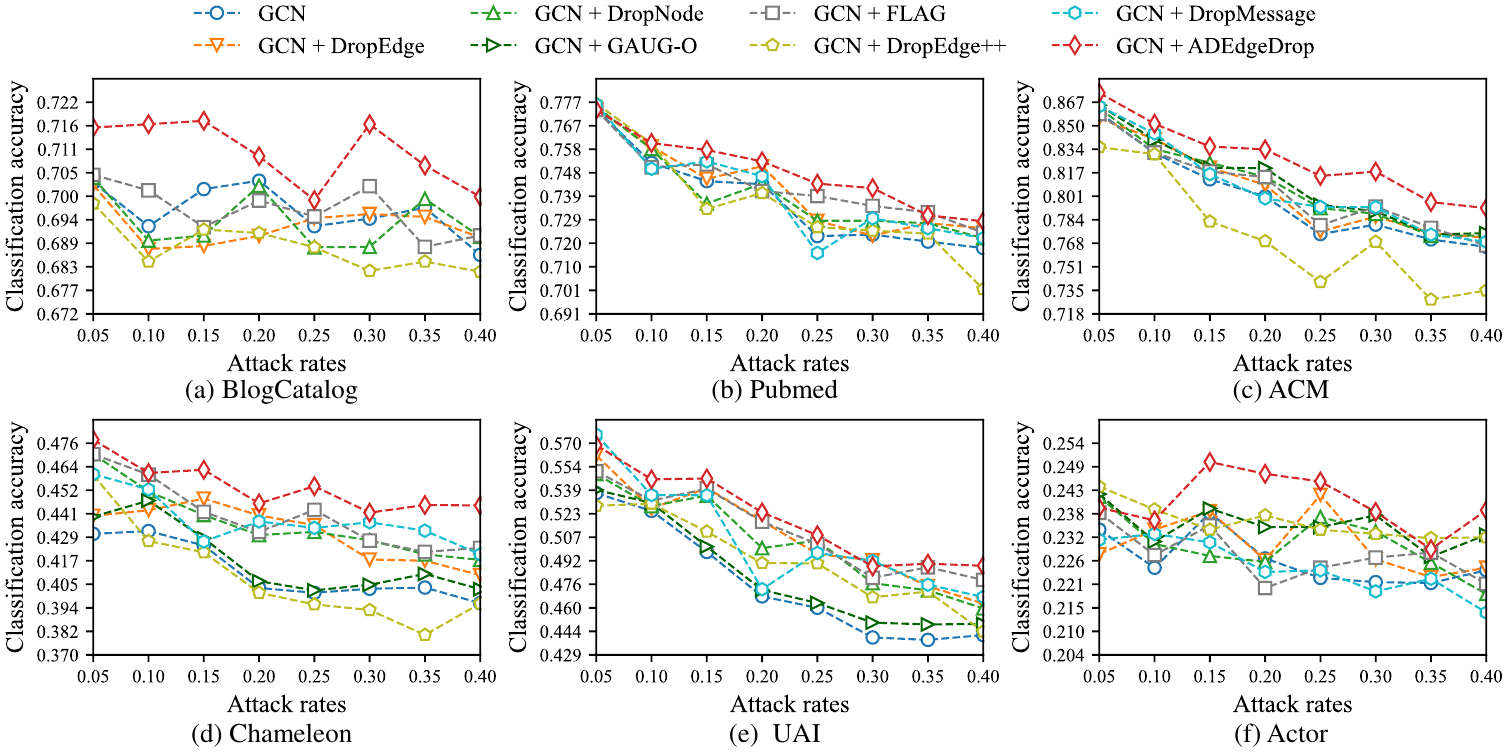}
    \caption{GCN classification accuracy of different GAL methods with varying rates of edge-addition attacks.}
    \label{EdgeAdd}     
  \end{figure*}

\begin{figure*}[t!]
    \centering
    \includegraphics[width=\textwidth]{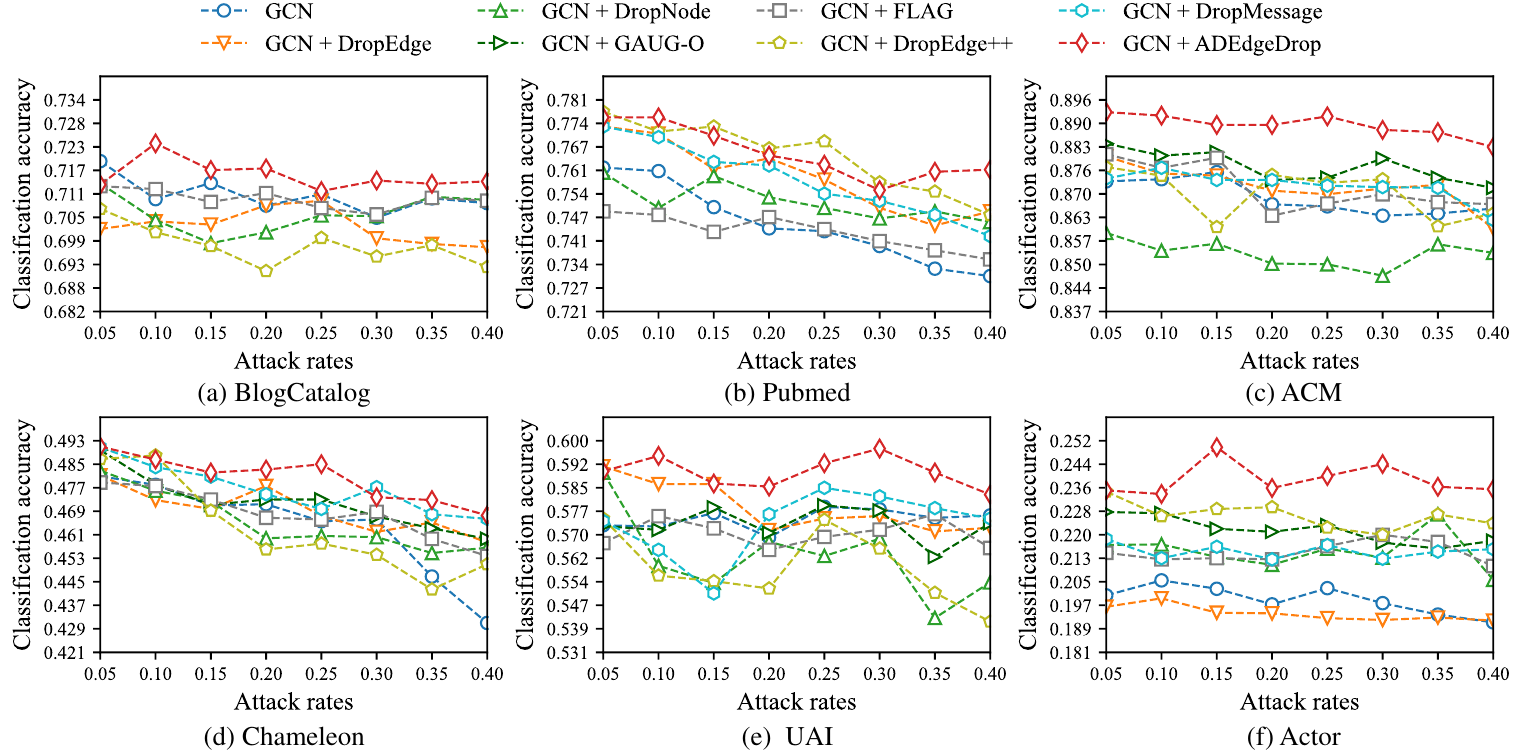}
    \caption{GCN classification accuracy of different GAL methods with varying rates of edge-removal attacks.}
    \label{EdgeDel}     
  \end{figure*}

In pursuit of testing the robustness of GNNs with the proposed method, we further conduct experiments on graphs with randomly added and removed edges.
In detail, we randomly remove or add $5\%$ to $40\%$ edges of the original graphs as attacks while ensuring that the new corrupted graphs are still undirected graphs.
Note that GAUG-O is not able to cope with the Pubmed dataset in an acceptable time owing to the high time complexity.
Figures~\ref{EdgeAdd} and~\ref{EdgeDel} show the GCN classification performance with different GAL strategies when encountering distinct types of edge-removal or edge-addition attacks with varied extents.
The experimental results indicate that ADEdgeDrop succeeds in alleviating the performance decline caused by edge-removal or edge-addition attacks.
Generally, the edge addition is more challenging to all GAL methods and it causes significant performance decline.
This may owe to the disrupted correspondence information between nodes by the newly added edges, while deleted edges have less influence since other multi-hop connections can often replace them.
In most cases, ADEdgeDrop achieves competitive accuracy when the graphs are corrupted,
demonstrating the effectiveness and robustness of our method when encountering edge-removal or edge-addition attacks.

\subsubsection{Analysis on Edge Dropping (RQ2)}

\begin{figure*}[!tbp]
    \centering
    \includegraphics[width=\textwidth]{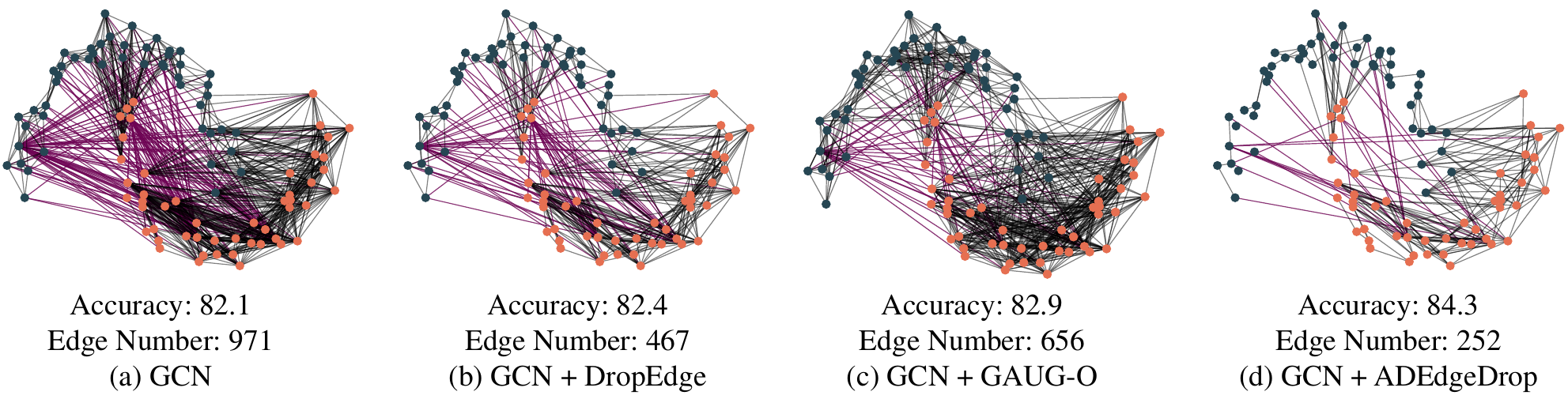}
    \caption{Visualization of retained edges generated by DropEdge, GAUG-O and ADEdgeDrop on the toy example dataset, where pink edges indicate noisy connections.}
    \label{VisToy}     
\end{figure*}

First, we generate a toy example to visualize the edge dropping results intuitively, as shown in Figure \ref{VisToy}.
There are 971 edges and 2 classes of nodes in this toy graph, where the pink edges indicate the noisy connections in the graph.
We visualize the graphs learned by the proposed ADEdegeDrop and other edge-based methods, from which we have the following observations.
We set the widely used drop rate for the random edge-dropping method, i.e., DropEdge, as 0.5.
It can be found that DropEdge gets a sparser graph with higher classification accuracy.
GAUG-O resamples the edges, thereby adding more connections and removing some unnecessary relationships simultaneously, leading to better performance.
As for the proposed ADEdgeDrop, our method significantly removes more noisy connections marked in pink.
It can be verified that ADEdgeDrop improves the classification results and obtains a high-quality graph with less noisy or redundant connections.
These experiments results also validate that most edges in a graph may be redundant, and the GNN training on an incomplete graph often leads to competitive performance.

In light of previous experimental analysis, a large number of edges in a graph are unnecessary or alternative.
With the message passing mechanism of GNNs, some edges can also be replaced by other multi-order paths. 
Thus, it is reasonable to learn a sparse and pure graph where important connections are retained and some redundant relationships are removed. 
Because the proposed model tends to preserve critical connections during training, we further evaluate GCN with randomly removed edges and the retained edges that are stored in the optimal training iteration of ADEdgeDrop.
In detail, we train a \emph{new basic GCN} with an incomplete graph learned by ADEdgeDrop.
We also generate another incomplete graph by randomly removing the same number of edges.
As reported in Table~\ref{SelectedPerf}, the percentage of deleted edges varies on distinct datasets because the optimal numbers of removed edges learned by ADEdgeDrop differ.
It is noted that the experimental results on graph $\hat{\mathcal{G}}_{AD}$ reported in Table~\ref{SelectedPerf} are not as high as the ADEdgeDrop model in Table~\ref{Performance}.
This is because we only use an incomplete graph for training in this experiment, while results reported in Table~\ref{Performance} utilize a complete graph for testing and only remove some edges when training and optimizing parameters.
However, compared with the performance of GCN with complete graph $\mathcal{G}$, GCN with the incomplete graph $\hat{\mathcal{G}}_{AD}$ generated by ADEdgeDrop still obtains comparable performance.
On BlogCatalog, ACM and Actor datasets, GCN with $\hat{\mathcal{G}}_{AD}$ even gains higher accuracy.
This indicates that the learned graph may have higher quality than the original graph by removing some noisy edges, and in most cases the backbone GCN can still achieve competitive performance with sparser graphs learned by ADEdgeDrop.
In addition, the quantitative results also reveal that the model training with graphs $\hat{\mathcal{G}}_{AD}$ learned by ADEdgeDrop gains superior accuracy than that training with $\hat{\mathcal{G}}_{RD}$ generated by random edge-dropping method,
thereby validating the feasibility of the incomplete graph learned by the proposed method.

\begin{table}[!tbp]
    \centering
    \caption{Percentage of deleted edges ($\varsigma$ \%) and GCN classification accuracy (\%) with complete graphs ($\mathcal{G}$),  incomplete graphs generated by the random edge dropping ($\hat{\mathcal{G}}_{RD}$) and ADEdgeDrop strategies ($\hat{\mathcal{G}}_{AD}$). The best accuracies are highlighted in \textbf{bold} and the second-best ones are \underline{underlined}.}
    \label{SelectedPerf}
    \begin{tabular}{l|c|ccc}
      \toprule
      Datasets    & $\varsigma$    &  GCN + $\mathcal{G}$     & GCN + $\hat{\mathcal{G}}_{RD}$ & GCN + $\hat{\mathcal{G}}_{AD}$ 
 \\\midrule
      BlogCatalog & 32.0  & \sperf{70.64} $\pm$ 1.44   & 70.13 $\pm$ 0.94     &  \bperf{70.68 $\pm$ 0.65}      \\
      Pubmed      & 56.9  & \bperf{77.14 $\pm$ 0.44}   & 72.22 $\pm$ 0.80      & \sperf{74.81 $\pm$ 0.71}     \\
      ACM         & 69.7  & \sperf{87.84 $\pm$ 0.19}   & 85.14 $\pm$ 1.00       & \bperf{88.93 $\pm$ 0.10}         \\
      Chameleon   & 49.6  & \bperf{48.21 $\pm$ 0.26}   & 44.92 $\pm$ 0.85       & \sperf{47.23 $\pm$ 0.40}  \\
      UAI         & 57.1  & \bperf{56.92 $\pm$ 0.67}   & 54.33 $\pm$ 0.31      & \sperf{56.64 $\pm$ 0.54}  \\
      Actor       & 22.0  & 20.76 $\pm$ 1.88   & \sperf{21.16 $\pm$ 1.39}       & \bperf{23.71 $\pm$ 0.93} \\
      CoraFull    & 35.0  & \bperf{56.94 $\pm$ 0.77}   & 54.62 $\pm$ 1.13       & \sperf{55.95 $\pm$ 0.99}  \\
      Arxiv       & 24.8  & \bperf{69.61 $\pm$ 0.15}   & 64.01 $\pm$ 0.17       & \sperf{68.22 $\pm$ 0.04}   \\\bottomrule
    \end{tabular}
\end{table}

\subsubsection{Impact of Threshold $\mu$ (RQ2)}

\begin{figure*}[!tbp]
\centering
\includegraphics[width=\textwidth]{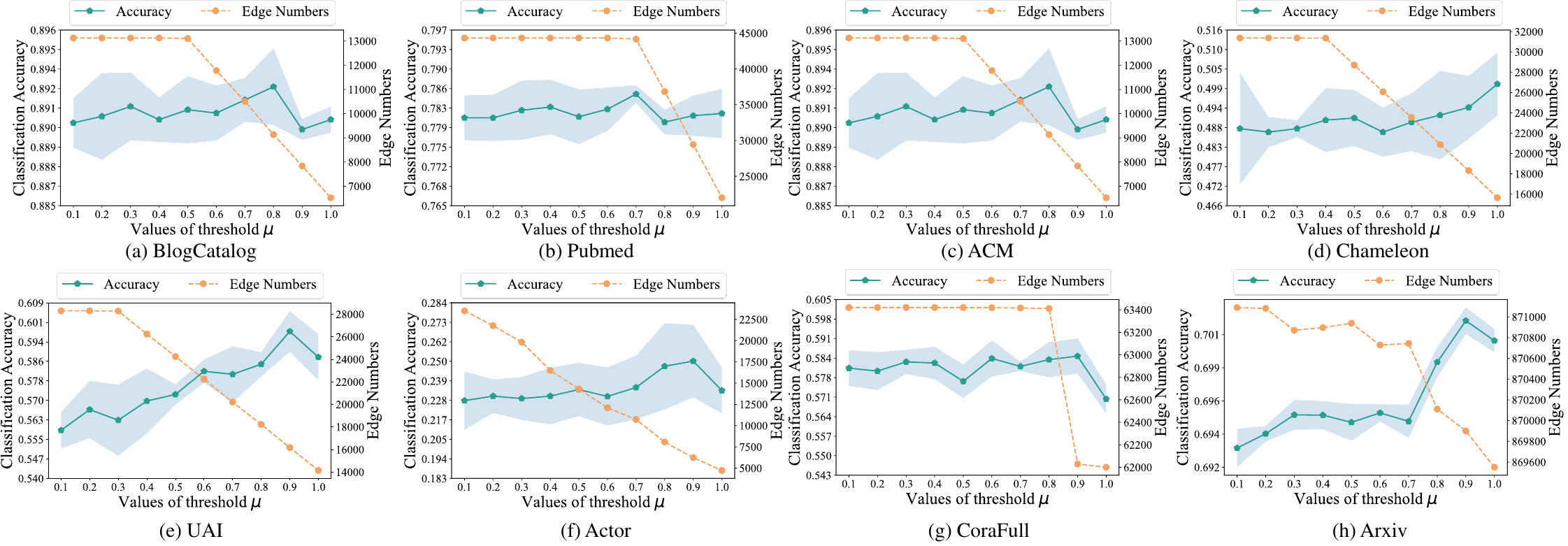}
\caption{Impact (test accuracy and edge numbers of the learned graph $\tilde{\mathcal{G}}$) of the threshold $\mu$ controlling the edge-dropping rate.}
\label{Thred}     
\end{figure*}

We next conduct the experiments to investigate the influence of the threshold $\mu$, which determines the number of removed edges during training.
We have the following observations from experimental results as demonstrated in Figure~\ref{Thred}.
First of all, we can find that more edges are deleted as the threshold $\mu$ increases.
Note that $\mu = 1$ does not necessarily mean that all edges have been removed due to the perturbation added to the output of the edge predictor.
Second, on most datasets, the optimal accuracy is achieved when $\mu>0.5$, indicating that a suitable deletion rate of edges is beneficial to the node representation learning.
On most datasets, it is significant that the accuracy tends to rise with the decreased number of edges until GNN reaches the highest accuracy.
After that, the classification accuracy may begin to fluctuate and dwindle, due to the excessive removal of edges in the graph.
This indicates that the remaining neighbor information is insufficient to achieve satisfactory performance.
In conclusion, it is critical to set a suitable threshold $\mu$ in different datasets, similar to the dropout rate of random dropping methods. 

  \begin{table}[!tbp]
    \centering
    \caption{Ablation study of ADEdgeDrop with GCN as the backbone. ADEdgeDrop w/o AT: ADEdgeDrop without the Adversarial Training. The best accuracies are highlighted in \textbf{bold}.}
    \label{AblationStudy}
    \begin{tabular}{l|cc}
      \toprule
      Datasets    & ADEdgeDrop w/o AT & ADEdgeDrop \\\midrule
      BlogCatalog & 71.94 $\pm$ 0.68       & \bperf{73.00 $\pm$ 0.68}    \\
      Pubmed      & 78.00 $\pm$ 0.71      & \bperf{78.52 $\pm$ 0.16}       \\
      ACM         & 88.82 $\pm$ 0.15      & \bperf{89.26 $\pm$ 0.08}       \\
      Chameleon   & 36.78 $\pm$ 3.50       & \bperf{50.06 $\pm$ 0.89}       \\
      UAI         & 57.62 $\pm$ 1.05       & \bperf{59.78 $\pm$ 0.81}       \\
      Actor       & 22.40 $\pm$ 1.45       & \bperf{23.04 $\pm$ 2.52}       \\
      CoraFull    & 57.50 $\pm$ 0.65     & \bperf{58.70 $\pm$ 0.46}       \\
      Arxiv       & 69.43 $\pm$ 0.13       & \bperf{70.19 $\pm$ 0.05}       \\\bottomrule
    \end{tabular}
  \end{table}

\subsubsection{Ablation Study and Effectiveness (RQ3)}
We further perform the ablation study to validate the effectiveness of the node similarity supervision, i.e., the effectiveness of the adversarial optimization target defined in Eq. \eqref{minmax}.
In light of this objective, we construct a simple edge predictor, which directly minimizes the training target $\mathcal{L}_{lg}$ defined in Eq. \eqref{recError}. 
Note that DropEdge in Table~\ref{Performance} can be regarded as a baseline without the edge predictor.
Table \ref{AblationStudy} reports the results of the ablation study, which reveals that ADEdgeDrop gains higher performance compared with that without the adversarial training.
The experimental observation points out that the node attribute similarities are important to the proposed strategy, and the adversarial optimization objective has the ability to utilize node affinity information to explore more reasonable edge-dropping schema.

% \subsubsection{Parameter Analysis (RQ 3).}
% We further look into the influences of hyperparameters in the proposed approach, including the threshold $\mu$, trade-off parameter $\alpha$ and updating rate $\gamma$ for the perturbation.

\subsubsection{Training Convergence Analysis (RQ3)}
\begin{figure*}[!tbp]
        \centering
        \includegraphics[width=\textwidth]{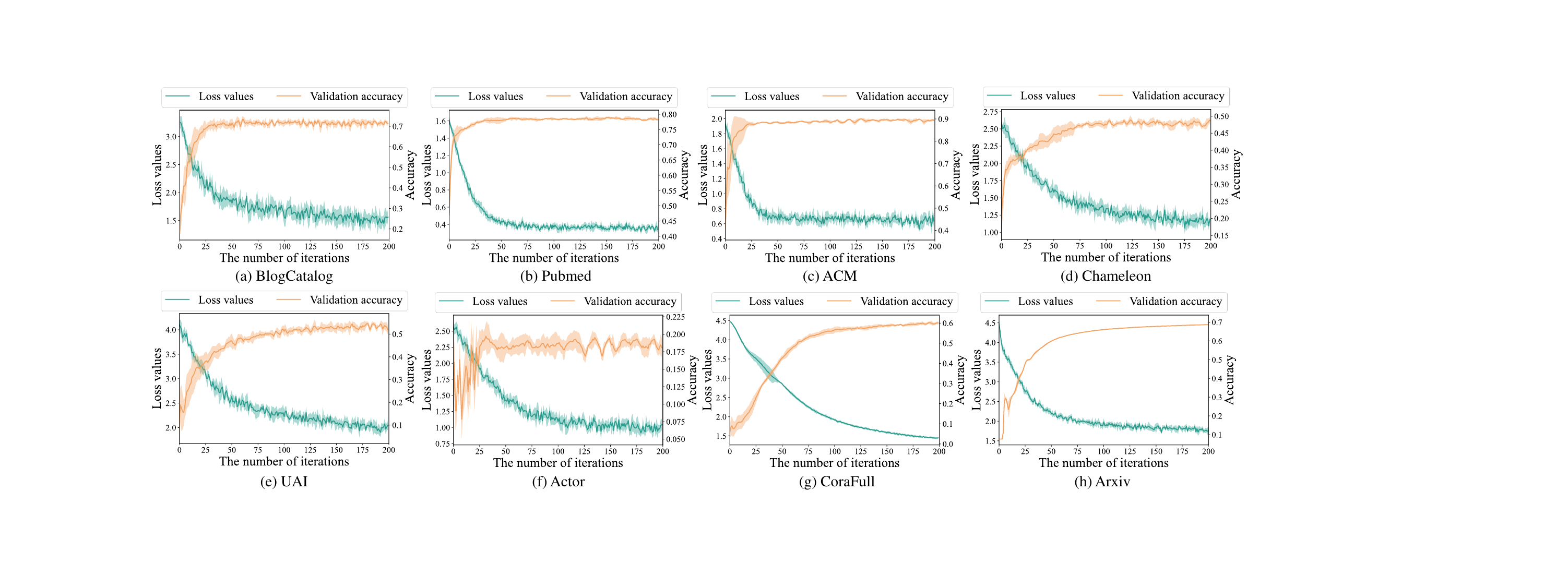}
        \caption{Curves of training losses and validation classification accuracy on tested graph datasets.}
        \label{Convergence}     
    \end{figure*}

We investigate the experimental convergence, as shown in Figure~\ref{Convergence}, which plots the curves of training losses and validation accuracy.
Owing to the perturbation added to the edge embedding, the graph structure constantly changes during training, thereby resulting in the fluctuation of training losses.
Nevertheless, the optimization algorithm of the model can lead to an overall decrease in objective losses and reach rough convergence.
The accuracy of the validation set also tends to achieve overall convergence with fluctuations.
These observations verify that the proposed multi-step training strategy of ADEdgeDrop can successfully minimize the training objectives.
  % \subsubsection{Parameter Analysis (RQ 3).}
  % We further look into the influences of hyperparameters in the proposed approach, including the threshold $\mu$, trade-off parameter $\alpha$ and updating rate $\gamma$ for the perturbation.
  
  \section{Conclusion}
  In this paper, we have proposed a novel edge-dropping strategy named ADEdgeDrop, which integrates the downstream GNNs with an adversarial edge predictor supervising the edge-dropping procedure in contrast to conventional random edge-dropping methods.
  We optimized the edge predictor by formulating a saddle point problem to seek the most vicious perturbations so that robust optimizations of both the edge predictor and the basic GNNs can be achieved.
  We guided training algorithms of perturbations and other weights by PGD and SGD to guarantee the full optimization of all trainable variables.
  Comprehensive experiments on eight benchmark datasets have demonstrated that the proposed approach outperforms the state-of-the-art GAL baselines.
  Moreover, ADEdgeDrop succeeded in improving the robustness of GNNs under different types of edge-removal or edge-addition attacks.
  In future work, we will further investigate robust GAL methods with the simultaneous addition and removal of edges.

\bibliographystyle{ieeetr}
\bibliography{MachineLearning}

\vfill

\end{document}